\documentclass[11pt]{article}

\usepackage[preprint]{acl}

\usepackage{times}
\usepackage{latexsym}

\usepackage[T1]{fontenc}

\usepackage[utf8]{inputenc}

\usepackage{microtype}

\usepackage{inconsolata}

\usepackage{graphicx}

\usepackage{booktabs}
\usepackage{adjustbox}
\usepackage{array}
\usepackage{amsmath}
\usepackage{graphicx}
\usepackage{amssymb}
\usepackage{bm}

\usepackage{xcolor}
\usepackage{tcolorbox}
\usepackage{hyperref}

\usepackage{todonotes}

\title{Authors Should Label Their Own Documents}

\author{
    Marcus Ma$^{*1,2}$ \quad
    Cole Johnson$^{*1}$ \quad
    Nolan Bridges$^{*1}$ \\
    \textbf{Jackson Trager}$^{2}$ \quad
    \textbf{Georgios Chochlakis}$^{2}$ \\
    \textbf{Shrikanth Narayanan}$^{2}$
    \\
    \\
    $^{1}$ \text{ }Echo Group \quad
    $^{2}$ \text{ } University of Southern California \\
    \texttt{\{marcus,cole,nolan\}@echogroup.ai}\\
    *\textit{ equal contribution}
}

\begin{document}
\maketitle

\begin{abstract}
Third-party annotation is the status quo for labeling text, but egocentric information such as sentiment and belief can at best only be approximated by a third-person proxy. We introduce author labeling, an annotation technique where the writer of the document itself annotates the data at the moment of creation. We collaborate with a commercial chatbot with over 20,000 users to deploy an author labeling annotation system. This system identifies task-relevant queries, generates on-the-fly labeling questions, and records authors' answers in real time. We train and deploy an online-learning model architecture for product recommendation with author-labeled data to improve performance. We train our model to minimize the prediction error on questions generated for a set of predetermined subjective beliefs using author-labeled responses. Our model achieves a 537\% improvement in click-through rate compared to an industry advertising baseline running concurrently. We then compare the quality and practicality of author labeling to three traditional annotation approaches for sentiment analysis and find author labeling to be higher quality, faster to acquire, and cheaper. These findings reinforce existing literature that annotations, especially for egocentric and subjective beliefs, are significantly higher quality when labeled by the author rather than a third party. To facilitate broader scientific adoption, we release an author labeling service for the research community at \href{https://academic.echogroup.ai}{academic.echogroup.ai}.

\end{abstract}
\section{Introduction}

Text alone cannot intuit intent. True language understanding requires contextualizing messages with communicators' source views shaped by individual preferences, beliefs, and judgments. However, when training language models to infer this context, access to the original source is often difficult or impossible. The prevailing method for interpreting source views is via observational proxy, whereby third-party annotators assign labels to text data based on their inferences of what the source author would have believed \cite{pustejovsky_2012}.

Third-party observers, however, are noisy proxies of the original source. Annotators lack ``privileged access'' \cite{vazire_2010} to authors' authentic internal states, meaning they lack situational context that can cause misinterpretations of original intent \cite{jones_1971, wallace_2014}. Decades of annotation research have documented fundamental limitations of third-party annotation, such as annotator biases, intent misalignment, and pervasive annotator disagreement \citep{snow_2008,fort_2016,aroyo_2015,uma_2021}. While some of these limitations can be mitigated, such as through inter-annotator analysis \cite{paletz_2024, chochlakis_2025}, these problems are structurally tied to third-party annotation. Reliance on third-party annotation is simply a logistical necessity, as it is impossible to reconstruct the original viewpoint during retroactive text labeling.

In this work, we investigate the technical feasibility of collecting annotation data directly from authors in real time. We present \textbf{author labeling,} a new annotation methodology where we present authors with the opportunity to annotate their documents at their moment of creation with LLM-generated questions. In the real-world setting of chatbot conversations, we deploy a data collection system that uses lightweight LLMs to automatically monitor user messages for relevance to various topics and tasks. When this system identifies task-relevant moments, we create a contextualized annotation task based on the user's query and present it to the user directly in the chatbot interface for them to answer.

\begin{figure*}[ht]
    \centering
    \includegraphics[width=\textwidth]{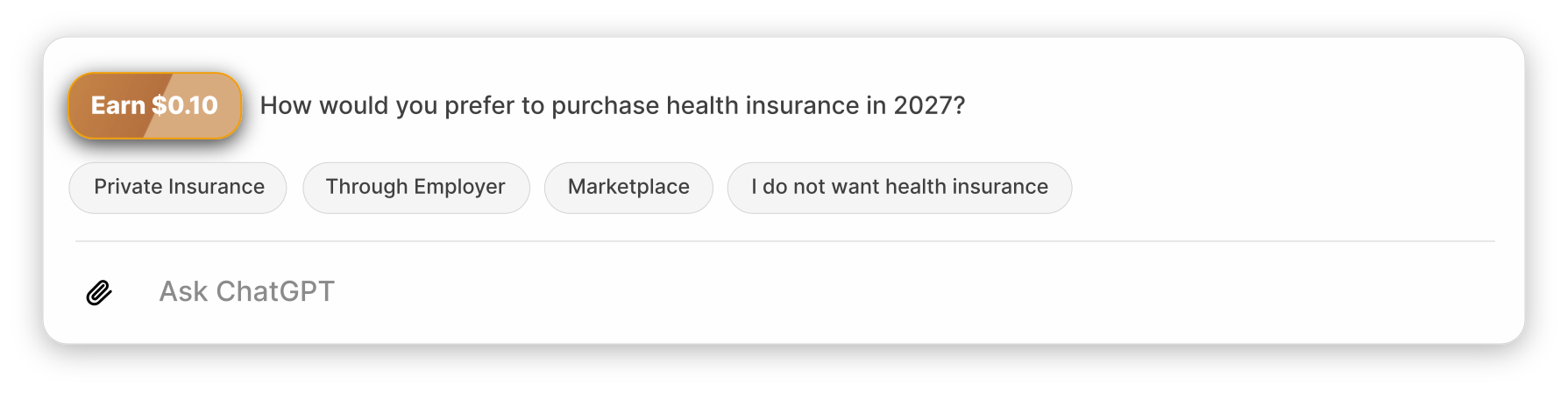}
    \caption{Example of an author-labeling task on echo for the query, ``want to switch to a new health insurance provider to cover dental''.}
    \label{fig:task_example}
\end{figure*}

To test author labeling, we introduce \textsc{Echo}, or \textbf{E}xperimental \textbf{C}oordination of \textbf{H}uman \textbf{O}bjectives, an online-learning model architecture that continuously improves through author labeling by strategically querying authors to reduce uncertainty over a predefined set of features or tasks. We implement \textsc{Echo} and deploy it into a live chatbot service with over 10,000 users for the task of product recommendation, a task that strongly depends on individual sentiment and preference with an objective measure of success, i.e. a click on the presented advertisement. We ran \textsc{Echo} side-by-side with an industry-standard banner-ad company's product and find \textsc{Echo} achieves a click-through rate (CTR) of 7.52\%, a fivefold increase over the industry baseline's 1.40\% CTR.  We also directly compare our author-labeling approach to other annotation methods for the task of sentiment analysis. Using three other annotation sources --- LLM-as-an-annotator, crowdsource via Mechanical Turk, and hand-selected expert --- we find author labeling is the most performant and internally consistent, and cheapest of the human methods. To support scientific adoption of author labeling, we release a public author labeling service at \href{https://academic.echogroup.ai}{academic.echogroup.ai}, enabling researchers to build datasets grounded in real-time author intent instead of third-party approximation.

\section{Author Labeling}
Author labeling provides the same theoretical benefits as other methods utilizing egocentric, real-time assessments such as patient-reported outcomes \cite{patrick_2011_content, FDA_PRO_2009} in the medical field and ecological momentary assessment \citep{shiffman_2008} in psychology. In \S\ref{subsec:motivation}, we highlight three advantages of author labeling over third-party annotation. In \S\ref{subsec:data_collection}, we provide an overview of how we conducted author labeling in a chatbot setting, and in \S\ref{subsec:limitations}, we discuss limitations of author labeling.

\subsection{Advantages of Author Labeling}
\label{subsec:motivation}

\paragraph{Intent Grounding.} Language is inherently ambiguous and can contain many valid interpretations. Skilled annotators can identify these ambiguities, but they have no means of resolving them. \citet{oprea_2020} find annotator performance on sarcasm detection --- an example of an ambiguous linguistic construct --- achieves an F1-score of 0.616 compared to authors' actual intents. Annotators often fail to consider the full background, personal context, and other non-linguistic influences that motivate the writing of the passage; \citet{buechel_2017} find instructions prompting annotators to explicitly consider the author's context and viewpoint increases label quality.

\paragraph{Ambiguity Resolution.} In subjective annotation tasks, disagreement between annotators can mean more than noise and can in some cases be a useful signal itself for detecting ambiguity \cite{beigman_2009, chochlakis_2025}. This ambiguity is simply a residue of the annotation process, though; language can be ambiguous to external annotators, but to the author, the meaning should be more clear\footnote{Whether meaning is clear to the original speaker is itself a contested belief \cite{nisbett_1977}. To this we assert meaning is at least \textit{more} apparent to the author than to a proxy.}.

\paragraph{Abstention as Signal.} Subjective annotation tasks see improved performance when providing annotators with an abstaining label such as ``not sure’’ or ``none of the above’’ \cite{uma_2022}. This hedges against ambiguous data where assigning a label would force annotators to guess, adding noise to the data. When authors themselves label their data with these choices, an abstention selection is no longer just a noise mitigation strategy but a valuable signal for model calibration and confidence. If authors were to label their own document with an abstention, then it is more likely to mean that there is no correct or applicable label to the document. When annotators label it so, they are simply lacking enough information to make a prediction on the author's internal beliefs.

\subsection{Data Collection}
\label{subsec:data_collection}

The largest obstacle for collecting data via author labeling is access to a diverse, large data source with an integrated UI platform for author annotations. We partner with Echo Group\footnote{https://chat.echogroup.ai}, a free-to-use chatbot platform that provides users with personalized product recommendations. Echo Group relays LLM inference to third-party LLM providers such as OpenAI and Anthropic and monitors the output stream. When recommendation for a product from Echo Group's catalog would benefit the user's conversation, it generates a conversation-contextualized and LLM-generated advertisement prompted to be as helpful and complementary to the chatbot response as possible.

We conducted author labeling with the purpose of gathering labeled information on a variety of intent features relevant to product recommendation, such as sentiment, urgency, and desired product features. We set up an automatic trigger that ran on every user query, where after a message was sent, we first passed the contents through a spam and greetings filter. We then evaluated the message with a lightweight LLM to see if it a given feature was relevant to the conversation. Some features, such as sentiment, were applicable to almost every query, while others, such as purchase urgency, were only relevant when the user was clearly exhibiting purchase intent. If enough intent features were deemed relevant to the user message, we presented users with a popup overlay asking them four multiple-choice or single-word free-text questions (Figure~\ref{fig:task_example}). These tasks are contextually generated on-the-fly, with each question designed to gain new information about its associated feature. We provide additional detail in \S\ref{subsec:features}.

\begin{figure*}
\centering

    \includegraphics[width=0.8\textwidth]{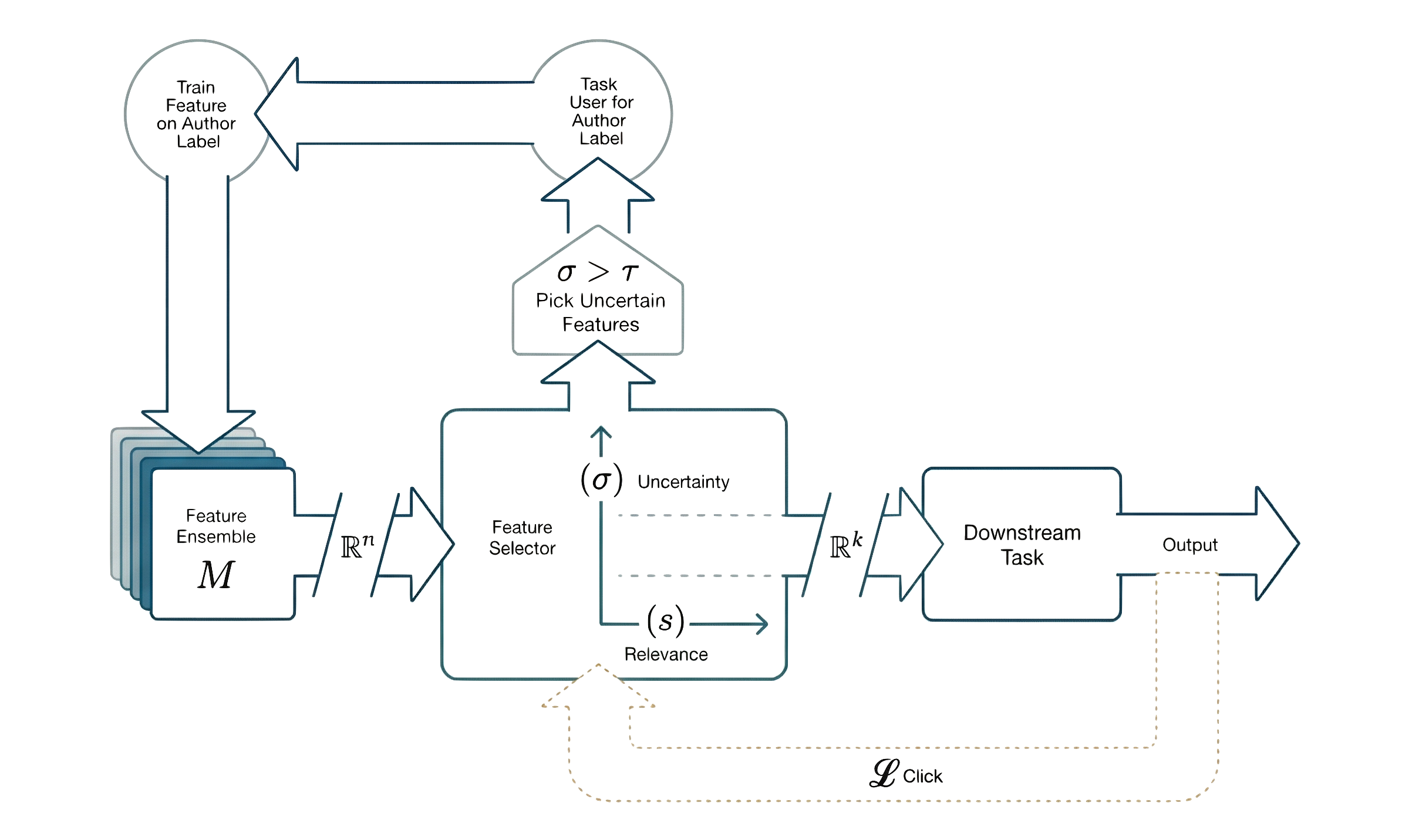}

    \caption{Diagram of the \textsc{Echo} architecture. Feature models are updated via author labeling tasks generated to minimize feature selector uncertainty. The feature selector is trained on losses from the downstream task.}
    \label{fig:echo}
    
\end{figure*}

\paragraph{Ensuring data quality.} We designed our label questions to be as non-intrusive and easy-to-answer as possible, drawing on established survey design principles that minimizing cognitive burden reduces the likelihood of low-quality or falsified data \cite{krosnick_1991, dillman_2014}; this is especially important given authors are not presented with detailed instructions typically given to third-party annotators. We designed all our questions to require either one-click or one-word answers and enforced a 5-second minimum reading time after presentation. We also relied on Echo Group's user fraud detection algorithms and only report data from trustworthy users (see \S\ref{app:spam}).

In October and November 2025, we processed 928,078 messages from 22,077 users. After filtering PII, sensitive topics, and spam, we displayed 54,977 four-question surveys and recieved 185,548 individual author-labeled datapoints, achieving a survey completion rate of 84.3\%.

\subsection{Limitations of Author Labeling}
\label{subsec:limitations}

Author labeling does not remove all ambiguity and noise from annotation. Different authors can have different interpretations of the same label, a well-known phenomenon for emotion and affect \cite{barrett2017emotions}. In self-reports, the phrasing of the question and potential labels influences the final answer significantly \cite{schwarz_1999}. The timing of data collection also greatly influences quality, regardless of the labeler, and even repeated self-reports from the same individual can have variance. Ephemeral details and post-hoc rationalization often change judgment from the original interpretation \cite{hoelzemann_2024}, but there is no definitive answer on which answer is most ``correct''. It is also well-known that inter-annotator variance adds a large source of variance for model performance and that simply adding or changing annotators can greatly change model behavior \cite{geva_2019}. Moving labeling to the authors themselves will not solve this issue for general systems, as each author still carries their own personalized viewpoint that can change over time. Additionally, author labeling does not solve demographic representation issues; however, when authors themselves annotate their data, authors' demographics become a feature to consider rather than a potential source of labeling bias. Observer bias and the Hawthorne effect \cite{landsberger1958hawthorne} can influence author answers, and some subjective tasks such as hate speech identification require a third party to determine if a statement is harmful. People also individually vary on their level of emotional intelligence \cite{salovey_1990} and ability to be aware of and distinguish their emotions \cite{barrett_1998, bagby_1994}.
\section{\textsc{Echo}}

To test author labeling, we introduce \textsc{Echo}, or \textbf{E}xperimental \textbf{C}oordination of \textbf{H}uman \textbf{O}bjectives, as an online-learning model architecture that continuously improves through author labeling by strategically querying authors to reduce uncertainty over a predefined set of features or tasks. \textsc{Echo} has three distinct phases: feature generation, feature selection, and the downstream task. Feature generation and feature selection are trained through two separate training loops using author labeling and downstream losses.

\begin{figure*}
    \centering
    \includegraphics[width=\textwidth]{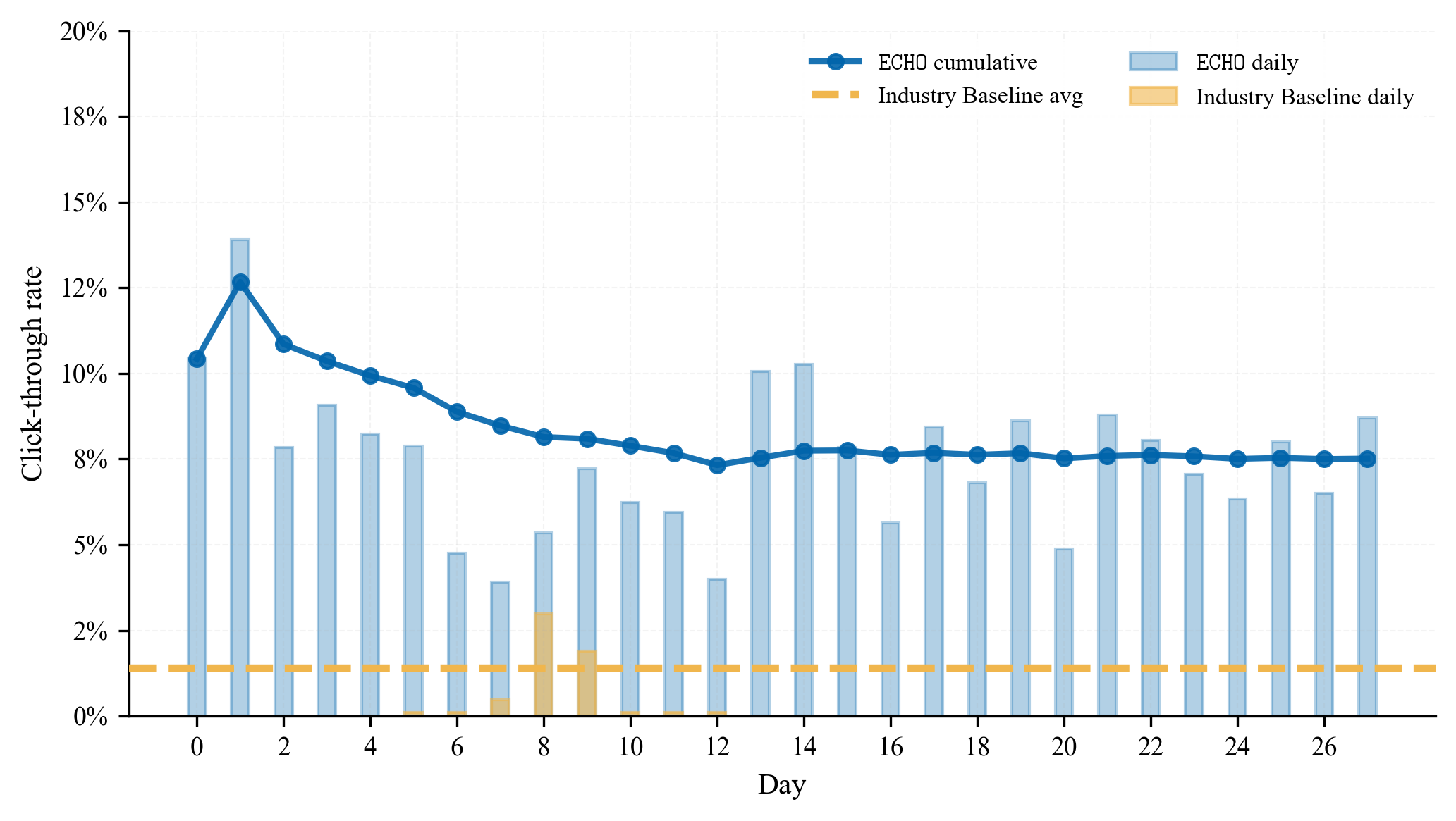}
    \caption{Click-through rate per day over the course of the 27-day experiment. The industry baseline was presented concurrently with \textsc{Echo} from days 5 to 12. Daily CTR for \textsc{Echo} and the baseline are shown as bars while \textsc{Echo}'s cumulative CTR is shown as the scatterline.}
    \label{fig:ctr_plot}
\end{figure*}

\subsection{Feature Generation}
\label{subsec:features}

We take a raw input $\mathbf{x}$ and create an ensemble of $n$ differentiable feature models, $\mathcal{M} = \{M_1, \dots, M_n\}$ parameterized by $\theta_{M_i}$. Each feature model is designed to predict one specific, distinct feature value (represented as a number, a class, a vector representation, etc.), meaning that we assume \textit{a priori} a set of all potential features that could be used by the downstream task. For a chatbot environment, for example, we could have $x$ to be the user message and $\mathcal{M}$ to be a collection of generative language models each tasked with identifying one specific linguistic or semantic aspect. We use each model to predict its relevant feature value:
\begin{equation}
    \hat{f}_i = M_i(\mathbf{x}; \theta_{M_i})
\end{equation}
and combine them into a candidate feature pool vector $\mathbf{\hat{F}}$.

\subsection{Feature Selector}
The feature selector, with parameters $\theta_S$, selects the top $k$ elements from $\mathbf{\hat{F}}$. This emphasizes the design decision that $n >> k$, for example having thousands of possible feature models and only picking the top ten for the downstream task\footnote{An alternative approach, if running all feature models on every input is expensive, is to train the feature selector to pick the top-k feature models $\{M_{s_1}, \dots, M_{s_k}\}$ rather than the top-k predicted features $\{\hat{f}_{s_1} \dots \hat{f}_{s_k}\}$.}.

The selector assigns a relevancy score $s_i \in \mathbb{R}$ and an uncertainty estimate $\sigma_i \in \mathbb{R}^+$ for each feature.
\begin{equation}
    [\mathbf{s}, \boldsymbol{\sigma}] = \text{Selector}(\mathbf{\hat{F}}; \theta_S)
    \label{eq:selector_output}
\end{equation}

We re-rank the features by their relevance score $s_i$ and pick the top-$k$ highest features, which we pass to the downstream model. The relevancy score $s$ is optimized via the downstream model's loss while the uncertainty $\sigma$ is updated via the author labeling response.

\subsection{Downstream Prediction}
We use the features passed from the feature selector in our downstream algorithm, which is configured to take exactly $k$ input features, which has a downstream evaluation and loss. In our implementation of a product recommendation system for example, described in further detail in \S\ref{sec:ctr}, the downstream recommendation algorithm receives specific user-intent features from the feature selector and produces the contextualized recommendation advertisement, where we use the click-through rate and more author labeling feedback (i.e., directly asking if the recommendation was helpful) as our downstream loss.

\subsection{Training}

We have two forms of feedback to supervise the training of \textsc{Echo}: the author labeling on specific feature predictions, which is used to train the feature models and the confidence of the feature selector; and the downstream loss, which trains the feature selector's relevance scores and optionally the downstream algorithm itself.

\subsubsection{Author Labeling}
After the feature selector identifies the top-$k$ features, we use the uncertainty scores $\boldsymbol{\sigma}$ and generate tasks to ask the user what the true value of that feature should be, making sure to include the predicted feature value as one of the options. When the author provides the ground-truth label for that feature, we use the value to update both the feature selector's uncertainty calculations and the associated feature model's prediction. We use this loss to update \textbf{only} the specific feature model, not all feature models, as the author labeling task is generated solely for that individual feature. To update the feature selector's uncertainty, we binarize the uncertainty loss into a 0 or 1 depending on if the label chosen by the author matches the feature model's predicted label. For free-response, loss is the cosine distance between predicted and author-labeled text embeddings.

\subsubsection{Downstream Loss}
We use the downstream loss to update the feature selector's feature selections; crucially, we stop backpropagation at the feature selector level. The downstream loss does not update the feature models themselves, since we do not know if the downstream loss was caused by bad predicted features or bad feature selection.
\section{Implementing \textsc{Echo} for Product Recommendation}
\label{sec:ctr}

\begin{figure}
    \centering
    \includegraphics[width=\linewidth]{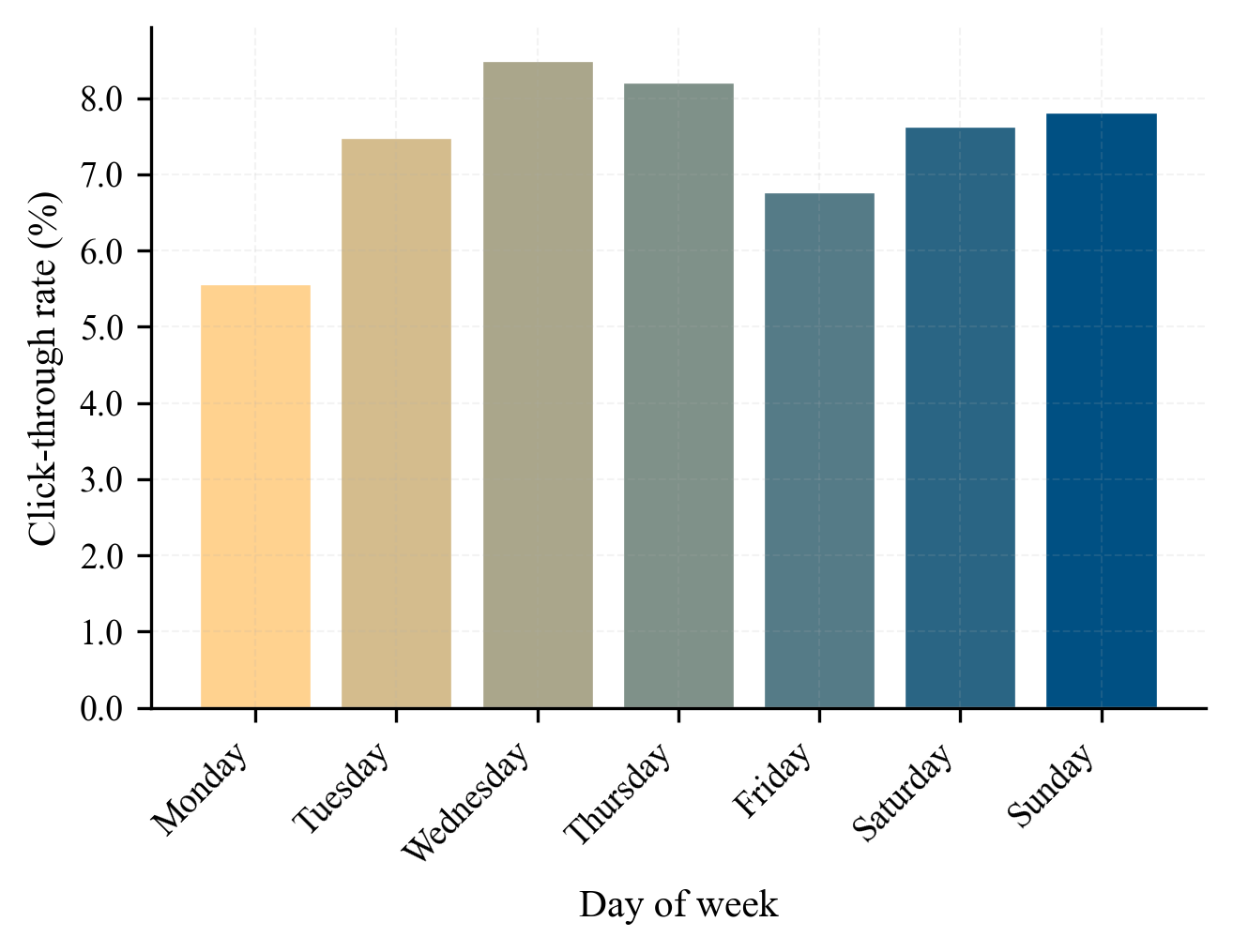}
    \caption{Day-by-day CTR Monday through Sunday.}
    \label{fig:day_ctr}
\end{figure}

We implement a version of \textsc{Echo} designed to recommend a product to the user based on their perceived needs from their conversation. Our feature models are lightweight LLMs prompted to guess the value of one of many features deemed potentially important for whether the user would be open to purchasing a product and also what specific product they would want, such as purchase timing and individual product specifications\footnote{We omit the full feature list, prompts, and model details to protect proprietary information from echo LLM.}. The feature label space differs per-feature, and we use a combination of binary values, free-form text, and a closed set of predefined labels depending on the feature.

The downstream algorithm is echo's product recommendation system, which takes in the selected features, matches it to the closest item in echo's product catalogue, and makes a final decision to display an advertisement rewritten to fit naturally into the conversation. Author labeling tasks are generated via an LLM call using a task generation template combined with feature specifications and definitions. We update our feature model LLMs with author labels through a combination of in-context learning, automated prompt engineering \cite{zhou_2023}, and supervised finetuning. We also update our downstream algorithm through author labeling, where we ask the user for feedback regarding the product recommendation itself.

\subsection{Experimental Setup}

We deploy our product recommendation implementation of \textsc{Echo} into a real-world chatbot environment. As the chatbot receives a user query, it passes the message quickly through \textsc{Echo}'s lightweight filter, which classifies messages as ``taskable'' or not (see Table~\ref{tab:annotation_examples} for examples). If the message is taskable, we pass the user query into the feature models and display author-labeling tasks for the selected features. We display these tasks independently of the downstream product recommender's decision to display an ad.

Before displaying any ads, we passed taskable messages through only phases 1 and 2 (feature generation and selection) with author annotation in order to train the feature models. We experimented with user-specific feature models but found limited success for individidual users with a lot of feedback, and decided ultimately on training feature models on all users' data to be more generalizable. We trained on about 50,000 author labels for about two weeks before launching the full version of \textsc{Echo} with product recommendation, beginning a 28-day experiment where we measured the click-through rate (CTR) of \textsc{Echo} compared to an industry baseline. Click-through rate is the ratio of user clicks to impressions, a standard measure in the advertisement industry for the effectiveness of our product recommendations. 

\paragraph{Industry Baseline}

During the four-week experiment, we also displayed advertisements provided by an established advertising company for an 8-day period. This advertiser presented banner ads, which are generated dynamically based on the user's query via their proprietary keyword-matching algorithm. During this period, we passed user queries to the industry baseline, and if it decided to service an ad, we displayed their advertisement in the same user interface as \textsc{Echo}'s recommendations.

\subsection{Results}

We plot the day-by-day CTR of \textsc{Echo} and the industry baseline in Figure~\ref{fig:ctr_plot}. In total, \textsc{Echo} created 10,378 impressions with 789 clicks for an overall CTR of 7.52\%, while the industry baseline presented 1,711 impressions with 24 clicks for a CTR of 1.40\%, a 537\% performance increase.

We also see a novelty bonus for \textsc{Echo}'s CTR: CTR spikes early into the introduction of each model then drops to a stable rate. This is consistent with advertisement literature finding that novelty perception for new stimuli decays rapidly and that increased exposure to a stimulus reduces advertisement engagement over time \cite{Wu_2007, Havlena_2004}.

While click-through rates for ad displays in chatbot environments have been understudied, the global average CTR for display ads on Google Display Network (i.e., on internet search) ranges between 0.46\% \citep{irvine_2024} and 0.57\% \citep{chaffey_2024}, with static banner ads seeing CTR as low as 0.10\% \cite{bannerflow_2024}. We believe the industry baseline's CTR of 1.40\% could be due to higher levels of attention and novelty for users using chatbots versus internet search. 

\paragraph{Day-by-day CTR variations.} We display day-by-day CTR in Figure~\ref{fig:day_ctr}. We observe Mondays and Fridays have the lowest average CTR with peaks mid-week and on the weekend, aligning with established consumer patterns. Monday's low performance aligns with the task-orientation online browsing hypothesis \cite{Bussiere_2016}, stating consumers allocate their cognitive resources to starting the work week, so buying products is not on their mind. Mid-week jumps could be explained by ``cyberloafing'' \cite{Lim_2002}, where online browsing increases as a distraction from work; the Friday dip could be best explained by higher rates of socialization, which is also evidenced by a dip in impressions on Friday (see Figure~\ref{fig:day_impressions} in Appendix~\ref{app:ctr}), with a weekend surge in CTR possibly explained by increased leisure and exploration from online consumers \cite{Moe_2004}.

\begin{figure}
    \centering
    \includegraphics[width=\linewidth]{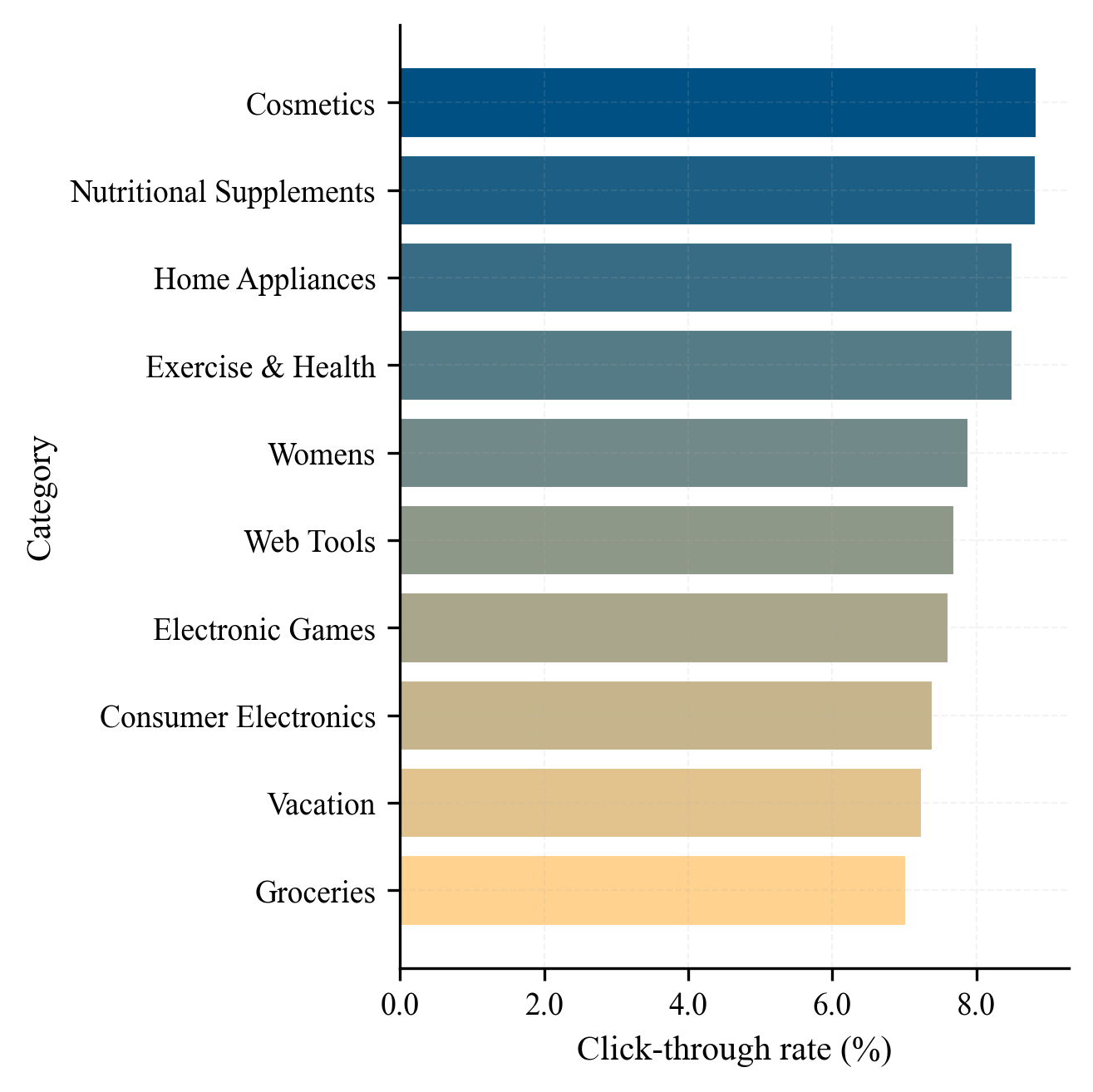}
    \caption{Top-ten verticals by CTR with a minimum of 180 impressions.}
    \label{fig:vertical_ctr}
\end{figure}
\section{Comparing Author Labeling to 3rd-Party Annotation}

We also perform a head-to-head analysis of author labeling for annotation compared to three status quo annotation methods. We use \textsc{Echo}'s multiple-choice questions generated for sentiment, where an LLM was probed to understand the user's emotional state and present four plausible emotions as options. We opted to use LLM-generated label options rather than predefined emotion labels so that the labels themselves could incorporate context for maximum user clarity (i.e. the label ``Concerned about pollution'' vs. ``Concerned''). We apply the same filtering criteria as \S\ref{subsec:data_collection} and restricted our sample to users who had completed a minimum of ten annotation tasks and sampled exactly six tasks for every user, applying a final manual inspection over all user messages and tasks. This yielded 1,962 total tasks across 327 users and 1043 conversations, with an average of 1.88 tasks per conversation and 3.19 conversations per user.

\subsection{3rd-Party Annotation Sources}

In addition to author labeling, we collect three types of annotation: LLM-as-an-annotator, crowdsourcing via Mechanical Turk (MTurk), and expert annotation. We chose these three methods to represent the status quo in annotation covering the high-volume scaling capacity of crowdsourcing \cite{snow_2008}, the emerging trends of using LLMs as low-cost annotators \cite{ding_2023}, and the quality control of expert annotation.

We report annotator demographics in Table~\ref{tab:demographics}; however, the authors have reason to believe that some of the demographics reported by Mechanical Turk annotators may have been falsified, as the authors received 15 emails from MTurk annotators with traditionally Indian or South Asian names written in non-fluent English yet only two MTurk annotators reported having an Asian ethnicity.

For the expert annotators, we selected a sample of 300 tasks from the 1,962 total. For LLM annotators, we used the same prompt for Llama 3.1 8B, Llama 3.3 70B, and GPT-OSS 20B as the three ``annotators''. We collected three annotations per datum for all annotation methods using 196 Mechanical Turk annotators, 30 expert annotators, and 5,886 LLM calls. We present additional details and prompts in Appendix~\ref{app:ann}.

\begin{table}
\centering
\small
\begin{adjustbox}{width=\linewidth}
\begin{tabular}{cccc}
\toprule
& \shortstack{Author\\Accuracy} & \shortstack{Annotator\\Agreement} & \shortstack{KL\\Divergence} \\
\midrule
LLM & \textbf{42.76\%} & \textbf{0.36} & \textbf{0.01}\\
MTurk & 33.18\% & 0.07 & 0.12\\
Expert & 42.33\% & 0.12 & 0.08\\
\bottomrule
\end{tabular}
\end{adjustbox}
\caption{Author accuracy, annotator agreement (Cohen's Kappa), and KL Divergence with the author label distribution for the three annotation methods.}
\label{tab:annotation_stats}
\end{table}

\subsection{Alignment with Author Labels}

We present overall accuracy with author labels, annotator agreement, and KL divergence with author label distributions for the three annotation methods in Table~\ref{tab:annotation_stats}\footnote{We note that random accuracy is 25\%, as all data are four-choice multiple-choice questions.}. We find experts (42.76\%) and LLMs (42.33\%) achieve essentially the same level of accuracy, but LLM annotators had much higher agreement ($\kappa$=0.36) than experts (0.12). However, we find that when experts do agree, their overall accuracy is higher (Table~\ref{tab:agreement_accuracy}), achieving the highest accuracy among methods of 63.0\% when all annotators agree on a label. Meanwhile, MTurk annotations perform the worst on all three metrics, which is to be expected compared to expert annotators but surprising in comparison to LLM annotation.

\subsection{Internal Consistency and Predictivity}

Alignment with author labels holds no meaning if the author labels themselves do not carry any useful signal. We conduct an experiment to see how \textbf{internally consistent and predictive} different annotation types are, including author labeling. Our goal is to use a user's prior answers to predict a future answer. In this setup, we prompt an LLM to predict what label the user would assign a question (Appendix~\ref{app:task_predictor}); however, we also provide as context \textbf{five other questions previously asked to the same user along with their labels} from five different sources: the author label, the three 3rd-party annotation methods, and random labels. Our intuition is that the predictor will perform better with the additional information given from other questions, and that the higher quality the label source, the better the performance.

\begin{table}
\centering
\small
\begin{adjustbox}{width=1.07\linewidth}
\begin{tabular}{cccc}
\toprule
& \shortstack{Three\\Agreements} & \shortstack{Two\\Agreements} & \shortstack{No\\Agreements} \\
\midrule
LLM & 50.5\% ($n=746$) & 39.6\% ($n=1027$) & 25.4\% ($n=189$)\\
MTurk & 37.4\% ($n=262$) & 35.1\% ($n=1223$) & \textbf{29.1\%} ($n=477$) \\
Expert & \textbf{63.0\%} ($n=46$) & \textbf{41.7\%} ($n=175$) & 25.3\% ($n=79$) \\
\bottomrule
\end{tabular}
\end{adjustbox}
\caption{Accuracy per annotation method based on number of agreements per datum. For ``No Agreements'', one of the three different answers is randomly selected.}
\label{tab:agreement_accuracy}
\end{table}

\paragraph{Results}

We present results in Table~\ref{tab:accuracy_comparison}. Performance follows the same pattern as in Table~\ref{tab:annotation_stats}: author labeling yields the highest accuracy, followed by LLM and expert annotators with comparable performance, and MTurk with the lowest overall accuracy. Because only the actual labels themselves changed between different prompts, we assert that the user labels are the most internally consistent and predictive of all methods tested.

\begin{figure*}
    \centering
    \includegraphics[width=\linewidth]{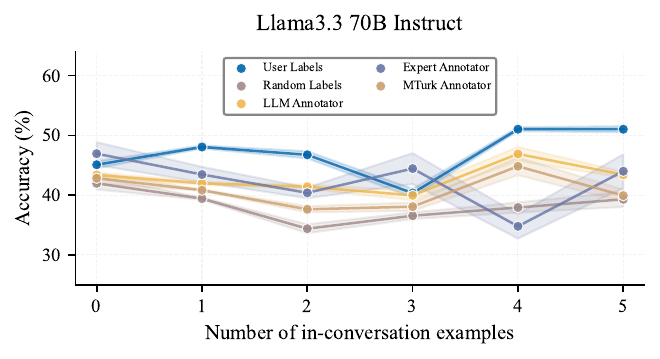}
    \caption{Predictor accuracy vs. number of in-conversation examples. All prompts always had 5 examples in total.}
    \label{fig:convo_analysis_llama_70b}
\end{figure*}

We also explored whether it mattered if the task examples occurred in the same conversation in Figure~\ref{fig:convo_analysis_llama_70b}. We find that some methods benefit from a higher number of in-conversation task answers while others do not, but also note that this differs by model as well (See Appendix~\ref{app:conv}). However, across all models, author labeling benefits the most from additional in-conversation examples, as intuitively additional context from the examples is likely most relevant in the same conversation and timespan of the actual question and author labeling contains the highest fidelity of annotator answer.

\begin{table*}[t]
\centering
\small
\begin{adjustbox}{width=0.8\textwidth}
\begin{tabular}{lcccc}
\toprule
Model & User Labels & LLM Annotator & Expert Annotator & MTurk Annotator \\
\midrule
Llama3.1 8B Instruct & \textbf{43.1$_{\pm 0.44}$} & 39.7$_{\pm 0.21}$ & 41.6$_{\pm 2.71}$ & 37.5$_{\pm 0.80}$ \\
GPT-OSS 20B & \textbf{46.2$_{\pm 0.13}$} & 41.5$_{\pm 0.16}$ & 40.4$_{\pm 3.34}$ & 40.2$_{\pm 1.13}$ \\
Llama3.3 70B Instruct & \textbf{46.9$_{\pm 0.52}$} & 42.7$_{\pm 0.68}$ & 43.6$_{\pm 2.46}$ & 40.8$_{\pm 0.54}$ \\
\bottomrule
\end{tabular}
\end{adjustbox}
\caption{Accuracy (\%) comparison across models and annotation methods. Values shown as mean$_{\pm \text{std}}$ over 3 runs. Best method per model shown in bold.}
\label{tab:accuracy_comparison}
\end{table*}

\subsection{Cost Analysis of Annotation Methods}

We conduct a cost and time analysis comparing author labeling to the two human annotation methods, Mechanical Turk crowdsourcing and expert annotation. We rewarded chatbot users with \$0.08 per task response, with each task taking less than ten seconds to complete (five seconds of forced response delay plus five seconds to answer). This yields a rate of \$28.80 per hour for author labeling tasks. For experts, while they graciously volunteered for free, we consider a rate of \$20 per hour to be reasonable. For Mechanical Turk, we offered a fixed rate of \$5 for a 30-question survey, which we initially estimated to take 30 minutes. In actuality, it took Mechanical Turk annotators an average of 9.1 minutes to complete the survey and expert annotators 17.0 minutes\footnote{The 9.1 minute survey completion for  MTurk likely indicates these annotators did not entirely read through each conversation, as assuming a fast reading rate of 250 WPM, this would mean they read 50 words per conversation, well below the 125-140 word average.}. We present the cost per datum and time per datum in Table~\ref{tab:cost_analysis}, which shows author labeling as significantly cheaper and faster than the other annotation methods. We note our calculations assume a single annotation per datum; these rates would double or triple when adopting conventional multiple-annotator coverage.

\begin{table}
\centering
\small
\begin{tabular}{cccc}
\toprule
& \shortstack{Cost\\per Datum} & \shortstack{Time\\per Datum} \\
\midrule
Author & \$0.08 & 10 sec\\
MTurk & \$0.17 & 18.2 sec \\
Expert & \$0.19 & 34 sec \\
\bottomrule
\end{tabular}
\caption{Time and cost analysis of author labeling, Mechanical Turk crowdsourcing, and expert annotation.}
\label{tab:cost_analysis}
\end{table}

\section{Conclusion}
Reliance on third-party annotation proxies for labeled text data arose out of technical necessity rather than quality improvement. In this work, we demonstrate a technical implementation of \textbf{author labeling} and ask document authors in real time to label their own data. As our two experiments show, author labeling is more internally consistent, cheaper, and faster than traditional 3rd-party annotation techniques and better retains the original author's signal and intent. We introduce a public author labeling service for the research community at \href{https://academic.echollm.io}{academic.echollm.io} to encourage adoption of author labeling as the new gold standard in annotation.
\section{Ethics}

In order to protect users’ privacy, we will not release any of the personal conversations used in our analysis besides the handful of non-identifiable examples used throughout this paper. We also redacted names, locations, and other PII when presenting it to our annotators and omitted any conversations dealing with sensitive topics such as personal health, religion, or finances. Users fully consented to their data being analyzed for business and scientific purposes when signing up to the chatbot and were fully informed on how they will be compensated for filing out label questions. Filling out label questions is always completely optional and a user's decision to fill out or not fill out a task does not affect their future experience on the platform in any way.

\section{Limitations}
\label{sec:limitations}

We previously discussed the theoretical limitations of author labeling in \S\ref{subsec:limitations}, but now discuss additional experimental limitations. Because of the real-world nature of our experiments, we could not control for all factors that occurred in the chatbot. The userbase itself was growing throughout the four-week trial period, and some spammy users were banned or excluded from analysis after the fact. Additionally, our head-to-head analysis of 3rd-party annotation to author labeling dealt with egocentric sentiment analysis. We did not analyze objective annotations or non-egocentric subjective tasks, both of which are tasks where being the author of the document might not be as important to annotation. In these scenarios, the qualifications of the author matter, which were factors that were outside the scope of this work. We also used LLMs to generate the questions in order to be as contextually relevant as possible, but the phrasing of questions can greatly influence self-reports \cite{schwarz_1999, abdurahman2024perils}. Additionally, although we introduce the general architectural framework of \textsc{Echo}, we only test its efficacy for the task of product recommendation.

\section{Acknowledgments}

The authors would like to thank all of the expert reviewers for volunteering to fill out our surveys. We would also like to thank Steve Wymer, Alex Vulakh, and other leadership at Echo Group.

\bibliography{custom}

\appendix
\onecolumn
\clearpage
\section{Additional Details on Product Recommendation Experiment}
\label{app:ctr}

\subsection{Impression Calculation}

We count each unique, individual instance of new content as a single impression, regardless of how many times it was displayed to the user (i.e., reloading when logging back in does NOT increase impression count). A single impression, if clicked multiple times, counts as multiple clicks, which is standard practice in the advertising industry. We noticed that some users would occasionally spam our chatbot with specific phrases with the hope of eliciting task rewards; we banned and excluded flagrant users from analysis, and to counteract additional spammy behavior, we merged impressions that occurred from messages that occurred too quickly in succession, as the only explanation for this user behavior would be for task spamming. We applied this calculation logic to both \textsc{Echo} and the industry baseline.

\subsection{Click and Impression Plots}

We present impression and click counts instead of CTR for figures \ref{fig:ctr_plot}, \ref{fig:day_ctr}, and \ref{fig:vertical_ctr} on the next pages.

\clearpage

\begin{figure*}
    \centering
    \includegraphics[width=\textwidth]{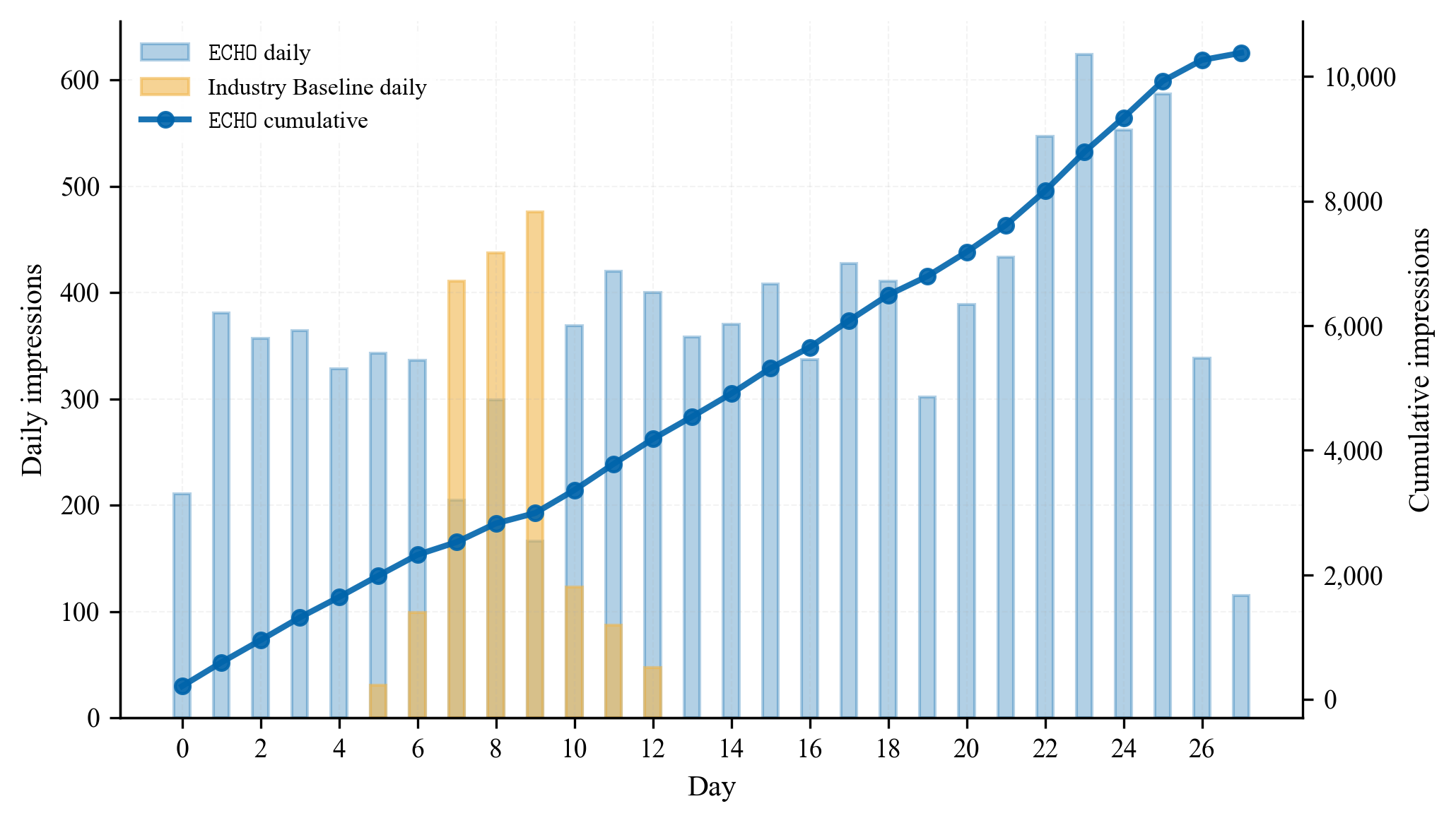}
    \caption{Impressions per day over the course of the 27-day experiment. The industry baseline was presented concurrently with \textsc{Echo} from days 5 to 12. Daily impressions for \textsc{Echo} and the baseline are shown as bars while \textsc{Echo}'s cumulative impressions are shown as the scatterline.}
    \label{fig:impressions_plot}
\end{figure*}

\begin{figure*}
    \centering
    \includegraphics[width=\textwidth]{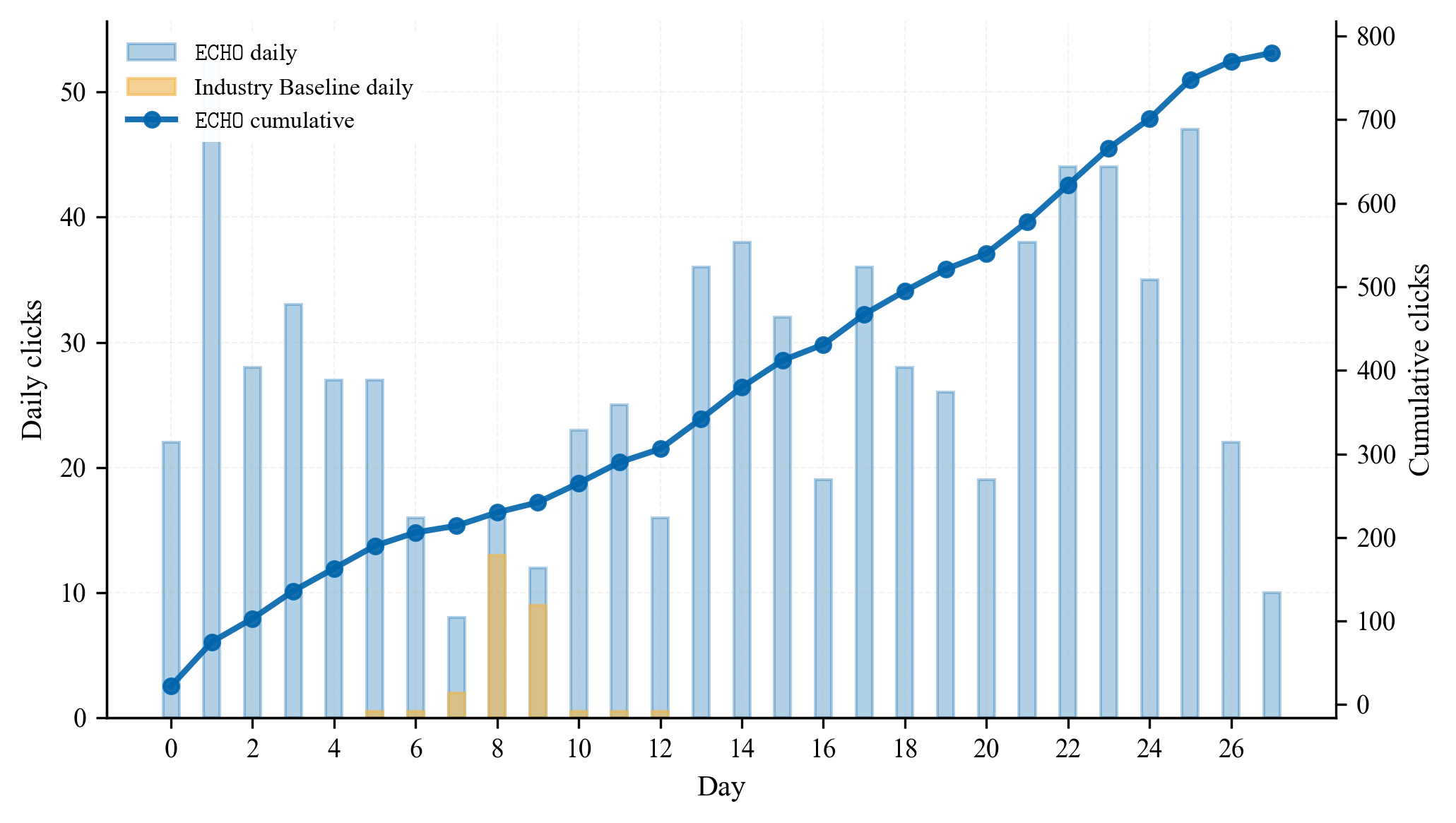}
    \caption{Clicks per day over the course of the 27-day experiment. The industry baseline was presented concurrently with \textsc{Echo} from days 5 to 12. Daily clicks for \textsc{Echo} and the baseline are shown as bars while \textsc{Echo}'s cumulative clicks are shown as the scatterline.}
    \label{fig:clicks_plot}
\end{figure*}

\clearpage

\begin{figure*}
    \centering
    \includegraphics[width=0.8\linewidth]{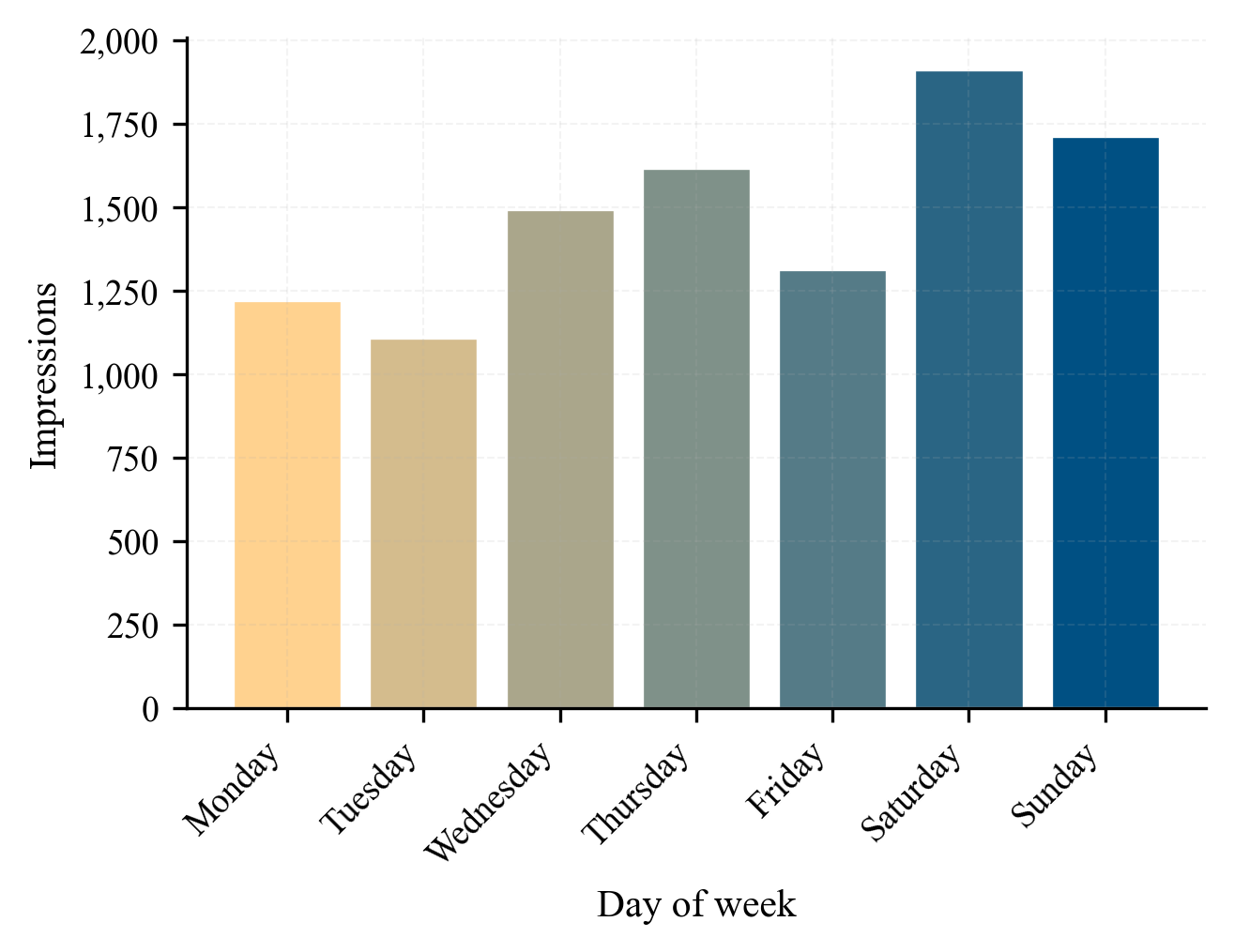}
    \caption{Day-by-day impressions Monday through Sunday.}
    \label{fig:day_impressions}
\end{figure*}

\begin{figure*}
    \centering
    \includegraphics[width=0.8\linewidth]{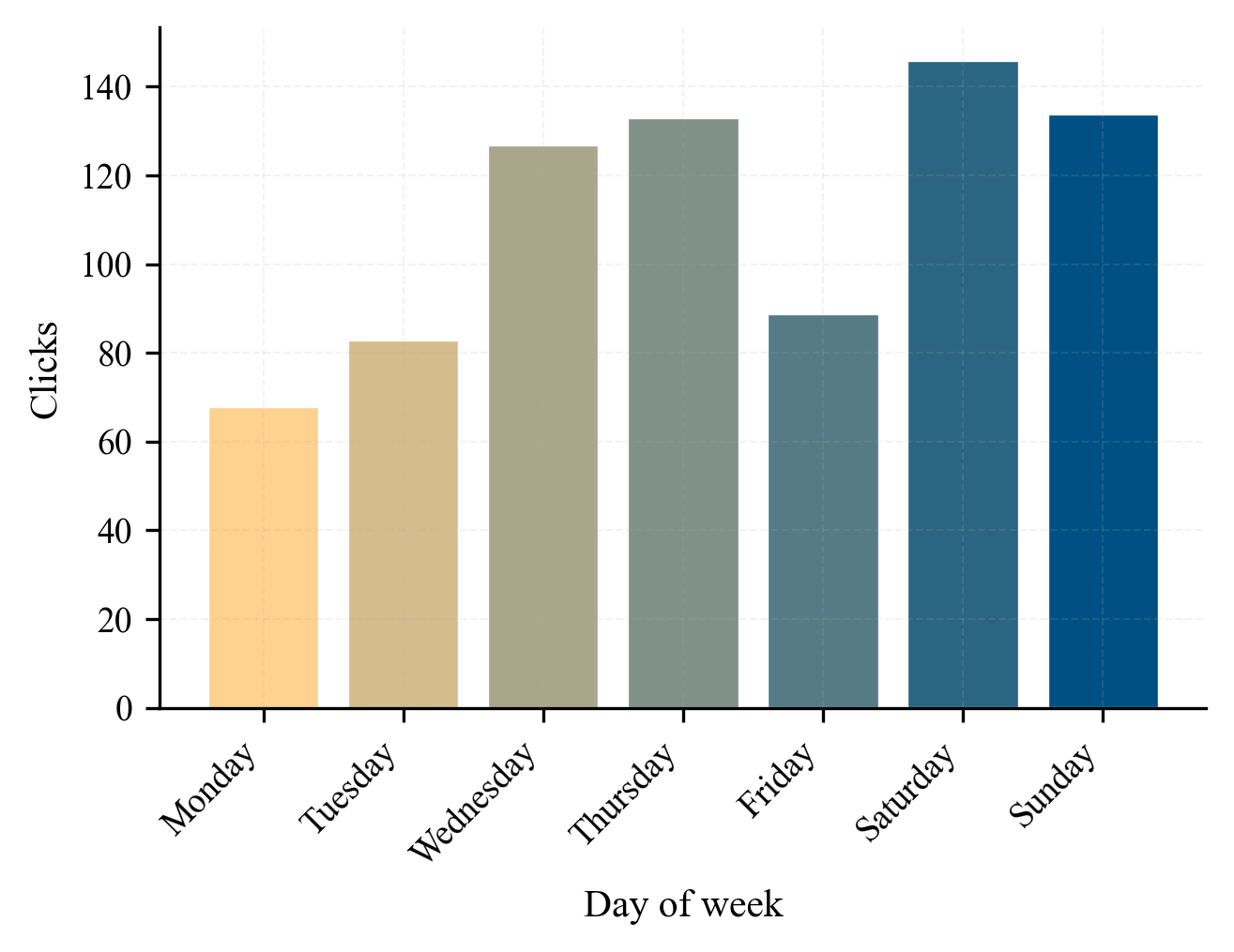}
    \caption{Day-by-day clicks Monday through Sunday.}
    \label{fig:day_clicks}
\end{figure*}

\clearpage

\begin{figure*}
    \centering
    \includegraphics[width=0.7\linewidth]{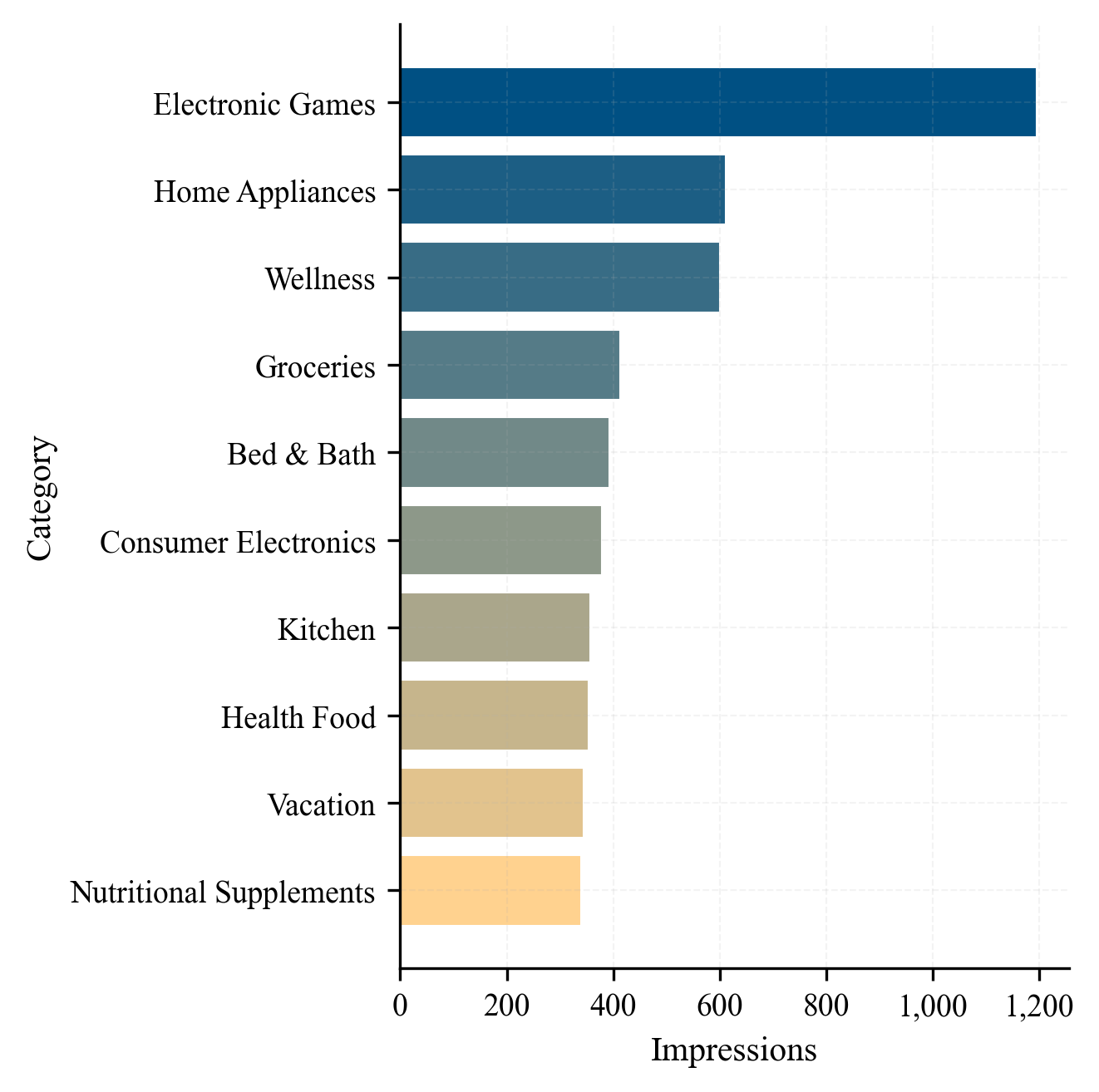}
    \caption{Top-ten verticals by impressions.}
    \label{fig:vertical_impressions}
\end{figure*}

\begin{figure*}
    \centering
    \includegraphics[width=0.7\linewidth]{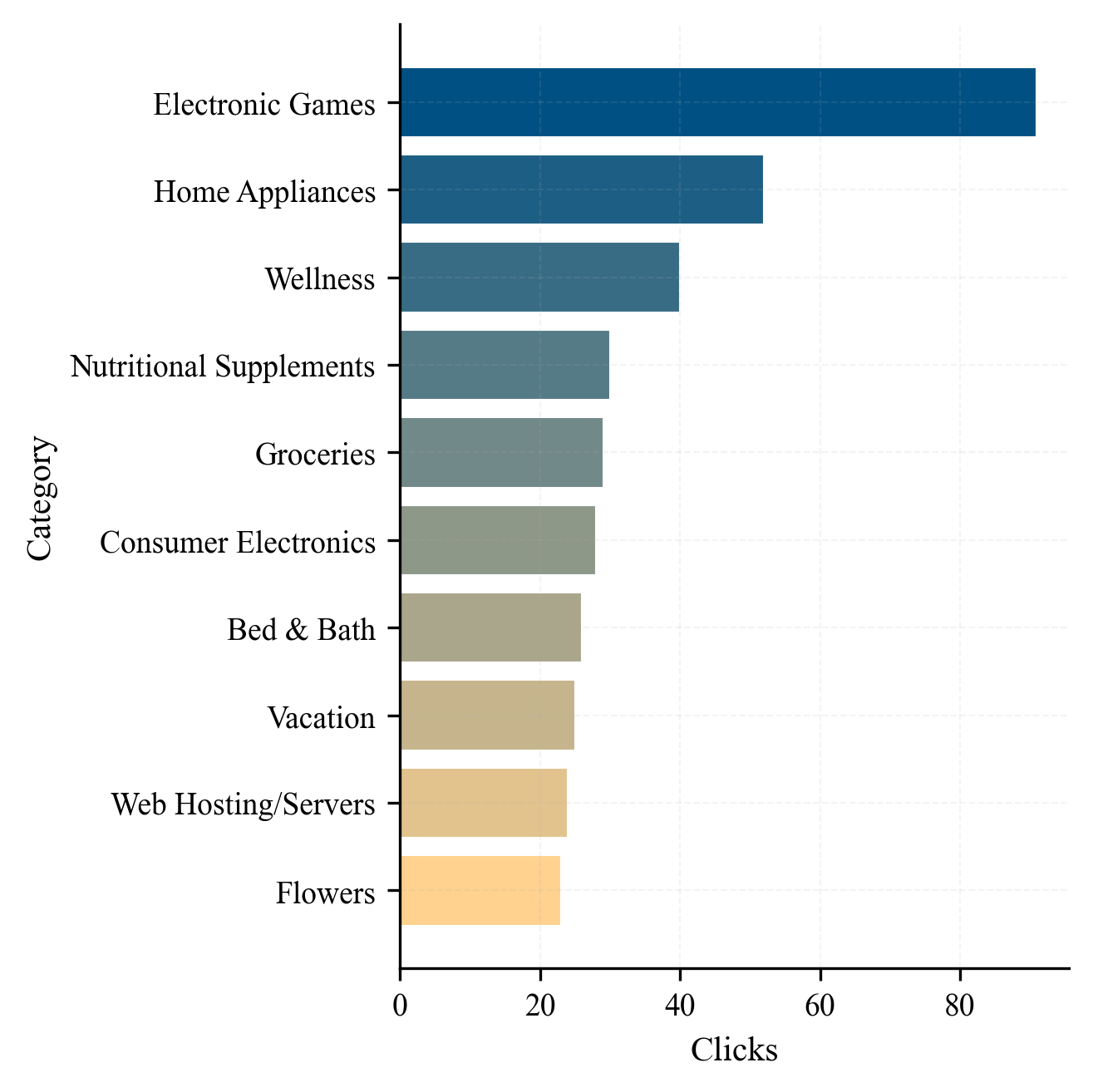}
    \caption{Top-ten verticals by clicks.}
    \label{fig:vertical_clicks}
\end{figure*}
\clearpage
\section{Annotation Details}
\label{app:ann}
\subsection{Qualtrics Survey}

We conduct all annotations via Qualtrics. Participants are informed that their participation is voluntary, confidential, and will be compensated (except for experts who graciously volunteered for free). The survey presented each task on its own page, first displaying the entire user conversation, then the question, then the answers. Long LLM responses were truncated to the first 300 characters while user messages were always retained in full. As a tradeoff between speed and complete contextual fidelity, we only displayed the five most recent messages of the conversation, which the authors acknowledge could in theory limit the total context of the conversation for very long conversations; however in practice, the vast majority of conversation context was unambiguously clear given the previous five messages.

\subsection{Non-US and Spam User Filtering}
\label{app:spam}

We filtered out several thousand users from analysis based on the following criteria:

\begin{enumerate}
    \item Location based outside the US
    \item Location inside the US, but using a known VPN location inside the US
    \item Location inside the US, but individual locations traversing the country several times a day, indicating VPN usage
    \item Consistent pattern of messages sent in order to just elicit tasks and their rewards
    \item Consistent pattern of messages that were spam or LLM-generated
\end{enumerate}

\subsection{Mechanical Turk Details}

We filter Mechanical Turk annotators to have over 95\% HIT approval rate and to have completed a minimum of 5000 HITs. We create 66 separate batches to annotate the 1,962 datapoints and have three unique annotators per datum. Annotation occurred entirely in November 2025.

\subsection{Expert Details}
We found expert annotators through snowball sampling. The authors had individual conversations with each expert on instructions for filling out the survey to ensure quality. The authors also personally verified all of the demographic information of each annotator. Annotation occurred from November 2025 to December 2025.

\clearpage
\subsection{Annotator Demographics}

\begin{table*}[htbp]
\centering
\begin{tabular}{lcc}
\toprule
\textbf{Characteristic} & \textbf{MTurk (N=196)} & \textbf{Expert (N=30)} \\
\midrule
\textbf{Age} & & \\
\quad Mean (SD) & 40.2 (10.3) & 27.8 (11.4) \\
\quad Range & 22--73 & 22--61 \\
\addlinespace
\textbf{Gender}, n (\%) & & \\
\quad Female & 90 (45.9) & 7 (23.3) \\
\quad Male & 105 (53.6) & 21 (70.0) \\
\quad Non-binary / third gender & 0 (0.0) & 1 (3.3) \\
\quad Prefer not to say & 0 (0.0) & 1 (3.3) \\
\addlinespace
\textbf{Race/Ethnicity}, n (\%) & & \\
\quad White & 186 (94.9) & 10 (33.3) \\
\quad Asian & 2 (1.0) & 12 (40.0) \\
\quad Black & 7 (3.6) & 1 (3.3) \\
\quad White/Asian & 0 (0.0) & 4 (13.3) \\
\quad White or European American,Hispanic or Latino/Latinx & 0 (0.0) & 1 (3.3) \\
\quad White or European American,Native Hawaiian or Pacific Islander & 0 (0.0) & 1 (3.3) \\
\quad Hispanic/Latinx & 1 (0.5) & 1 (3.3) \\
\addlinespace
\textbf{Education}, n (\%) & & \\
\quad High school graduate & 4 (2.0) & 1 (3.3) \\
\quad Some college & 7 (3.6) & 0 (0.0) \\
\quad 2 year degree & 11 (5.6) & 0 (0.0) \\
\quad 4 year degree & 135 (68.9) & 21 (70.0) \\
\quad Professional degree & 32 (16.3) & 2 (6.7) \\
\quad Doctorate & 5 (2.6) & 6 (20.0) \\
\addlinespace
\textbf{Political Party}, n (\%) & & \\
\quad Democrat & 87 (44.4) & 21 (70.0) \\
\quad Republican & 91 (46.4) & 3 (10.0) \\
\quad Independent & 18 (9.2) & 5 (16.7) \\
\addlinespace
\textbf{Political Ideology}, n (\%) & & \\
\quad Very liberal & 4 (2.0) & 3 (10.0) \\
\quad Liberal & 12 (6.1) & 6 (20.0) \\
\quad Slightly liberal & 5 (2.6) & 6 (20.0) \\
\quad Moderate & 11 (5.6) & 9 (30.0) \\
\quad Slightly conservative & 24 (12.2) & 2 (6.7) \\
\quad Conservative & 108 (55.1) & 3 (10.0) \\
\quad Very conservative & 29 (14.8) & 1 (3.3) \\
\bottomrule
\end{tabular}
\caption{Demographics of Mechanical Turk and Expert Annotators.}
\label{tab:demographics}
\end{table*}

\clearpage
\subsection{LLM Annotator Prompt}

We used the following prompt to act as our LLM annotator. We used Llama 3.1 8B Instruct, Llama 3.3 70B Instruct, and GPT OSS 20B as our three ``annotators''. Text in blue corresponds to the conversation, red corresponds to the question, and green corresponds to the task options that are substituted into each prompt.

\begin{tcolorbox}[
    colback=gray!5!white, 
    colframe=gray!50!black, 
    title=\textbf{Prompt Template for LLM Annotator},
    fontupper=\small\ttfamily, 
    boxrule=0.8pt,
    arc=2mm
]
You are a human subjects annotator tasked with guessing the answer to a sentiment-related multiple choice question from the perspective of a user. You will be provided with the user's conversation, which you should use to inform your answer.

\vspace{0.5em}
--BEGIN CONVERSATION--\\
USER QUERY: \textcolor{blue}{Will coffee make the hair smell like coffee and for how long until the smell disappears?}\\
------------------------\\
CHATBOT RESPONSE: \textcolor{blue}{Yes, using coffee as a hair dye can leave a subtle coffee scent in your hair. The smell may linger for a few hours to a couple of days, depending on your hair type and how well you rinse it out. If the scent is strong initially, it should fade as your hair dries and with subsequent washes. Using a m...}\\
------------------------\\
USER QUERY: \textcolor{blue}{Will coffee dye leave a stain on clothing after it has set into the hair?}\\
--END CONVERSATION--

\vspace{0.5em}
Now, from the user's perspective, how would they answer this question?

\vspace{0.5em}
--BEGIN QUESTION--\\
\textcolor{red}{How do you feel about the potential coffee smell in your hair after using a coffee dye?}\\
--END QUESTION--

\vspace{0.5em}
--BEGIN OPTIONS--\\
\textcolor{olive}{A. Frustrated\\
B. A bit annoyed\\
C. Neutral\\
D. Wanting to avoid it
}\\
--END OPTIONS--

\vspace{0.5em}
Respond with exactly one character: A, B, C, or D.\\
Do NOT add explanations, punctuation, quotes, or any other text.

\vspace{0.5em}
Respond ONLY with one of these four exact strings:\\
A\\
B\\
C\\
D

\vspace{0.5em}
Choose one and output ONLY that line, with no additional characters or text.
\end{tcolorbox}
\clearpage
\section{Additional Details on Sentiment Analysis Experiment}
\label{app:sentiment}

\subsection{Dataset Details}

\begin{figure}[h]
    \centering
    \includegraphics[width=0.5\linewidth]{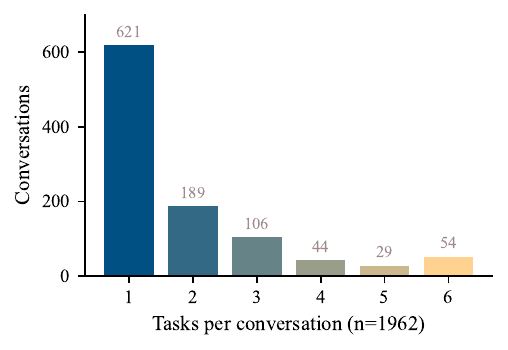}
    \caption{Distribution of the 1,962 sampled sentiment tasks across conversation.}
    \label{fig:task_per_conv}
\end{figure}

We present the number of tasks per conversation in Figure~\ref{fig:task_per_conv} and example tasks in Table~\ref{tab:annotation_examples}.

\begin{table}[h]
\centering
\small
\begin{tabular}{>{\centering\arraybackslash}m{3.50cm}>{\centering\arraybackslash}m{3.00cm}>{\centering\arraybackslash}m{3.00cm}}
\toprule
\midrule
User Message & Task Question & Options \\
\midrule
How to make a blanket? & What motivated you to want to make a blanket? & \shortstack{A. To relax and unwind \\ B. As a gift for someone \\ C. For personal use \\ D. Out of curiosity} \\ \midrule
Are there any 96 degree indoor pools around Panama City, FL & What's driving your need for a 96-degree pool in Panama City? & \shortstack{A. Looking to relax \\ B. Need stress relief \\ C. Just curious \\ D. Seeking comfort}  \\ \midrule
How to start an online business & How do you feel about starting your online business? & \shortstack{A. Excited and motivated \\ B. Nervous about the risks \\ C. Unsure where to begin \\ D. Determined to succeed}  \\ \midrule
\bottomrule
\end{tabular}
\caption{Sample of three taskable messges and the sentiment analysis task question generated for it. Conversation context is not shown here but was to annotators.}
\label{tab:annotation_examples}
\end{table}

\clearpage

\subsection{Prompt for Task Predictor}
\label{app:task_predictor}

We used the following prompt to act as our task predictor. Text in blue corresponds to the conversation, red corresponds to the question, purple corresponds to examples and answers, and green corresponds to the task options that are substituted into each prompt.

\begin{tcolorbox}[
    colback=gray!5!white, 
    colframe=gray!50!black, 
    title=\textbf{Prompt Template for LLM Annotator},
    fontupper=\small\ttfamily, 
    boxrule=0.8pt,
    arc=2mm,
    width=\dimexpr\linewidth+2cm\relax, 
    enlarge left by=-1cm,               
    enlarge right by=-1cm               
]
You are a human subjects annotator tasked with guessing the answer to a sentiment-related multiple choice question from the perspective of a user. You will be provided with the user's conversation, which you should use to inform your answer.

\vspace{0.5em}
--BEGIN CONVERSATION--\\
USER QUERY: \textcolor{blue}{Will coffee make the hair smell like coffee and for how long until the smell disappears?}\\
------------------------\\
CHATBOT RESPONSE: \textcolor{blue}{Yes, using coffee as a hair dye can leave a subtle coffee scent in your hair. The smell may linger for a few hours to a couple of days, depending on your hair type and how well you rinse it out. If the scent is strong initially, it should fade as your hair dries and with subsequent washes. Using a m...}\\
------------------------\\
USER QUERY: \textcolor{blue}{Will coffee dye leave a stain on clothing after it has set into the hair?}\\
--END CONVERSATION--

\vspace{0.5em}

The user has also filled out other sentiment-related multiple-choice questions.

\vspace{0.5em}

These are answers to questions from the same conversation:\\
\textcolor{purple}{
MESSAGE: is it easy to make a DIY natural dye?\\
QUESTION: How do you feel about the uncertainty of DIY natural dye duration?\\
ANSWER: Concerned}

\vspace{0.5em}

These are answers to questions from different conversations:\\
\textcolor{purple}{
MESSAGE: How do Airbnb stays compare to boutique hotels for long-term travel?\\
QUESTION: How do you feel about the uncertainty of long-term travel accommodations?\\
ANSWER: Anxious about the unknown\\ \\
MESSAGE: no, i want the tiktok styles\\
QUESTION: What’s the main emotion driving your school meal rant?\\
ANSWER: Frustration\\ \\
MESSAGE: tell me more about it\\
QUESTION: What's driving your desire to learn more about swimming?\\
ANSWER: Curiosity\\ \\
MESSAGE: Is Hulu cheap?\\
QUESTION: How do you feel about the cost of streaming services like Hulu?\\
ANSWER: I'm looking for good value
}

\vspace{0.5em}
If applicable, you should use inferred or explicit information about the user's emotional state from these prior answers, and factor in whether they came from the same or different conversation as this question. If they are not applicable or unrelated, then you do NOT need to incorporate their answers in your final decision.

\vspace{0.5em}
Now, from the user's perspective, how would they answer this question?

\vspace{0.5em}
--BEGIN QUESTION--\\
\textcolor{red}{How do you feel about the potential coffee smell in your hair after using a coffee dye?}\\
--END QUESTION--

\vspace{0.5em}
--BEGIN OPTIONS--\\
\textcolor{olive}{A. Frustrated\\
B. A bit annoyed\\
C. Neutral\\
D. Wanting to avoid it
}\\
--END OPTIONS--

\vspace{0.5em}
Respond with exactly one character: A, B, C, or D.\\
Do NOT add explanations, punctuation, quotes, or any other text.

\vspace{0.5em}
Respond ONLY with one of these four exact strings:\\
A\\
B\\
C\\
D

\vspace{0.5em}
Choose one and output ONLY that line, with no additional characters or text.
\end{tcolorbox}

\clearpage

\subsection{Predictor Accuracy vs. In-Conversation Examples for Additional Models}
\label{app:conv}

\begin{figure*}[h] 
    \centering
    
    \includegraphics[width=0.95\linewidth]{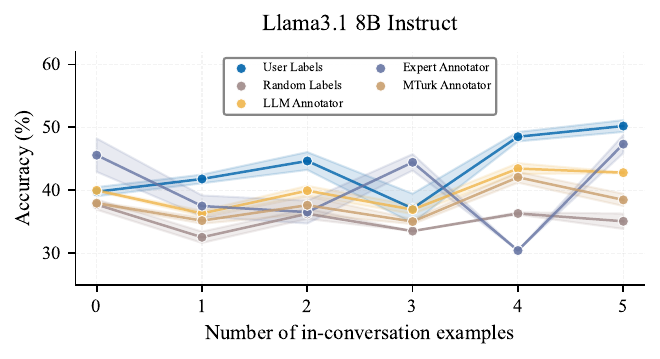}
    \caption{Predictor accuracy vs. number of in-conversation examples for Llama 3.1 8B Instruct.}
    \label{fig:convo_analysis_llama_8b} 
    
    \vspace{0.5cm} 

    \includegraphics[width=0.95\linewidth]{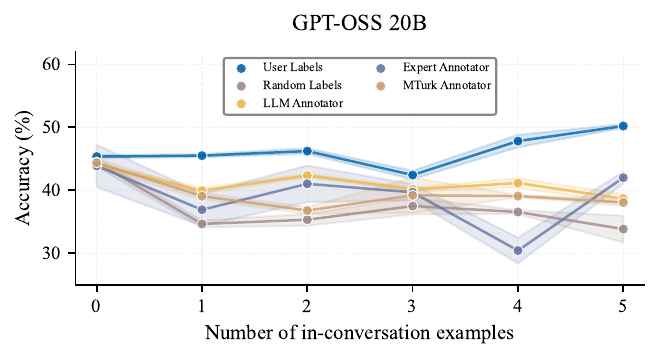}
    \caption{Predictor accuracy vs. number of in-conversation examples for GPT OSS 20B.}
    \label{fig:convo_analysis_gpt_oss} 

\end{figure*}

\clearpage

\subsection{Predictor Accuracy vs. Conversation Length}

\begin{figure*}[ht!] 
    \centering
    
    \includegraphics[width=0.6\linewidth]{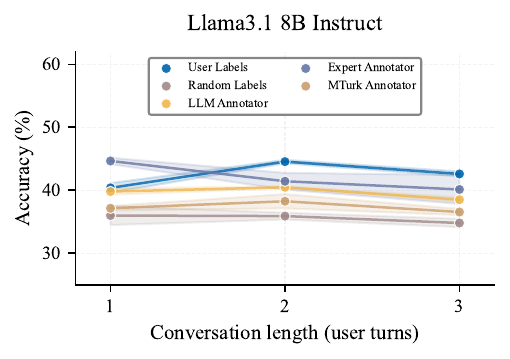}
    \caption{Predictor accuracy vs. conversation length for Llama 3.1 8B Instruct.}
    
    \vspace{0.5cm} 

    \includegraphics[width=0.6\linewidth]{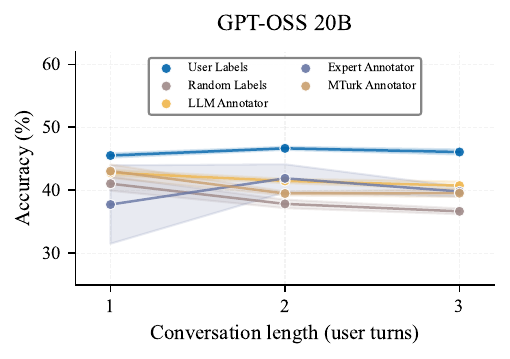}
    \caption{Predictor accuracy vs. conversation length for GPT OSS 20B.}

    \includegraphics[width=0.6\linewidth]{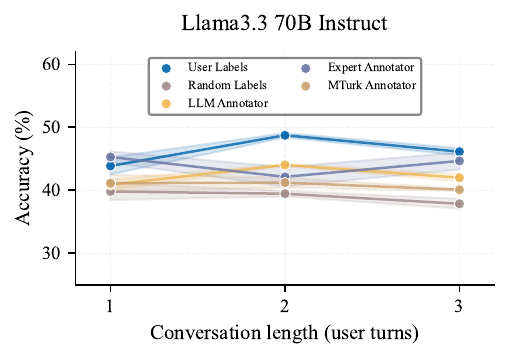}
    \caption{Predictor accuracy vs. conversation length for Llama 3.3 70B Instruct.}

\end{figure*}

\end{document}